\documentclass[final]{cvpr}

\usepackage{times}
\usepackage{epsfig}
\usepackage{booktabs}
\usepackage{graphicx}
\usepackage{amsmath}
\usepackage{amssymb}

\usepackage{color}
\usepackage{colortbl}
\usepackage{pifont}
\usepackage{subfigure}
\usepackage{multirow}

\usepackage{lipsum}

\newcommand\blfootnote[1]{%
  \begingroup
  \renewcommand\thefootnote{}\footnote{#1}%
  \addtocounter{footnote}{-1}%
  \endgroup
}

\definecolor{red}{rgb}{1,0,0}
\definecolor{blue}{rgb}{0,0,1}

 \makeatletter
 \def\hlinewd#1{%
      \noalign{\ifnum0=`}\fi\hrule \@height #1 \futurelet
      \reserved@a\@xhline}

\usepackage[pagebackref=true,breaklinks=true,colorlinks,bookmarks=false]{hyperref}

\def\cvprPaperID{8058} 
\def\confYear{CVPR 2021}

\begin{document}

\title{Bridge to Answer: Structure-aware Graph Interaction Network \\ for Video Question Answering
}

\author{Jungin Park \quad\quad\quad\quad Jiyoung Lee \quad\quad\quad\quad Kwanghoon Sohn\thanks{Corresponding author.} \\ Yonsei University, Seoul, South Korea \\ {\tt\small $\lbrace$ newrun, easy00, khsohn $\rbrace$@yonsei.ac.kr}}


\maketitle

\begin{abstract}
    \blfootnote{This research was supported by the Yonsei University Research Fund of 2021 (2021-22-0001).}
    This paper presents a novel method, termed Bridge to Answer, to infer correct answers for questions about a given video by leveraging adequate graph interactions of heterogeneous crossmodal graphs.
    To realize this, we learn question conditioned visual graphs by exploiting the relation between video and question to enable each visual node using question-to-visual interactions to encompass both visual and linguistic cues.
    In addition, we propose bridged visual-to-visual interactions to incorporate two complementary visual information on appearance and motion by placing the question graph as an intermediate bridge.
    This bridged architecture allows reliable message passing through compositional semantics of the question to generate an appropriate answer.
    As a result, our method can learn the question conditioned visual representations attributed to appearance and motion that show powerful capability for video question answering.
    Extensive experiments prove that the proposed method provides effective and superior performance than state-of-the-art methods on several benchmarks.
\end{abstract}\vspace{-10pt}

\section{Introduction}
    Video question answering (VideoQA) is a task to answer the question regarding a given video in a natural language form.
    Over the past few years, several methods have been focused on manipulating spatio-temporal visual representations conditioned by linguistic cues for VideoQA~\cite{tvqa, song18, movieqa, wang18}.
    However, because of its specificities such as dynamic spatiotemporal dependencies of the video and sophisticated compositional semantics of the question, the VideoQA still remains a challenging problem.
    
    \begin{figure}[t]
    \begin{center}
    \renewcommand{\thesubfigure}{}\hfill
       \subfigure[(a) Example for VideoQA]{\includegraphics[width=1\linewidth]{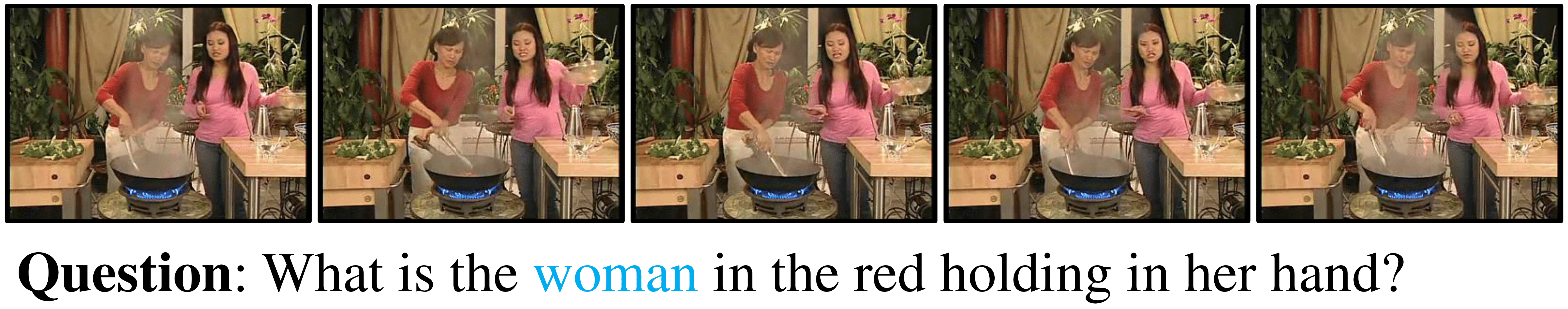}}\hfill     \\
       \subfigure[(b) Q2V interactions]{\includegraphics[width=0.49\linewidth]{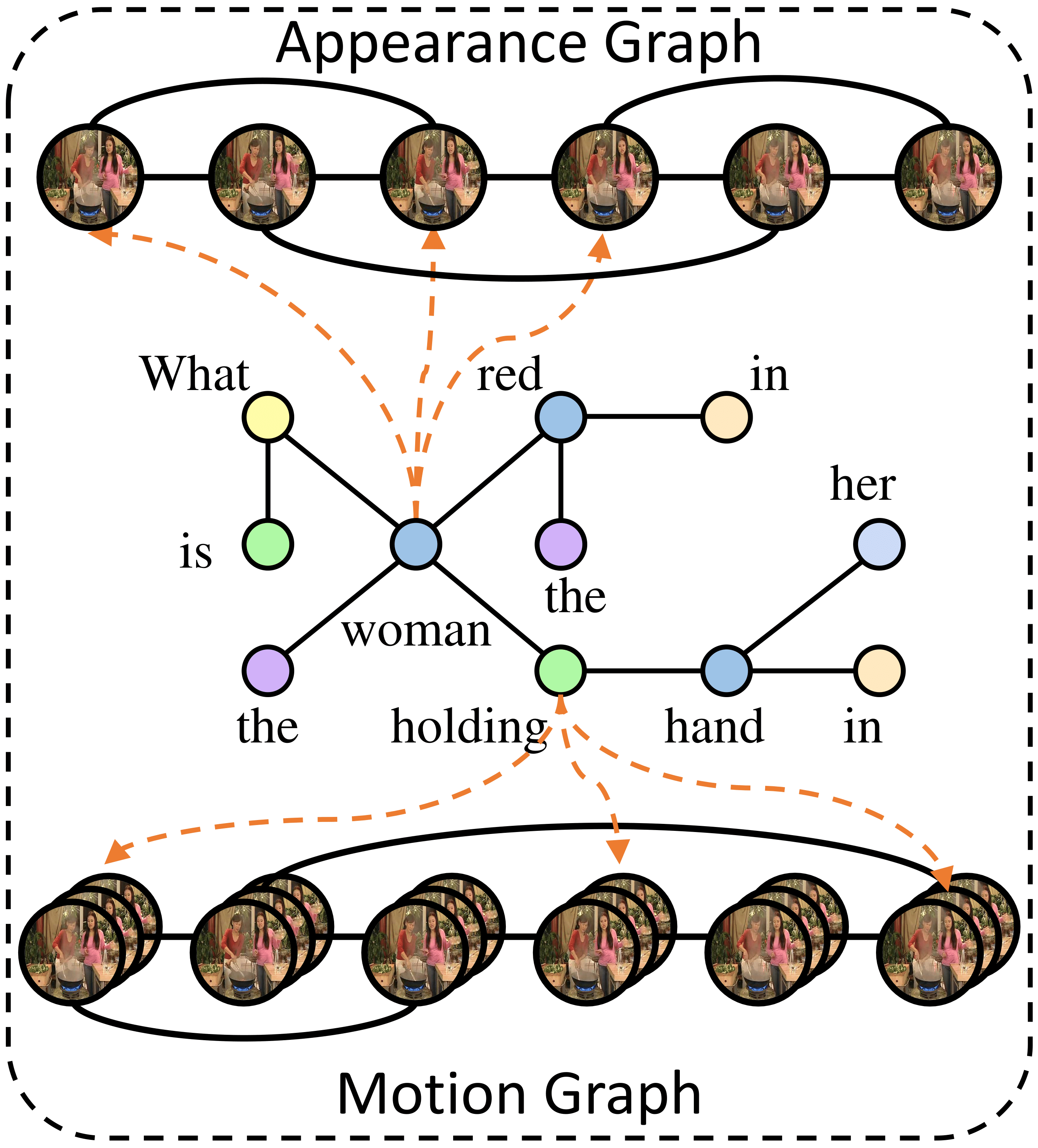}}\hfill
       \subfigure[(c) M2A interactions]{\includegraphics[width=0.49\linewidth]{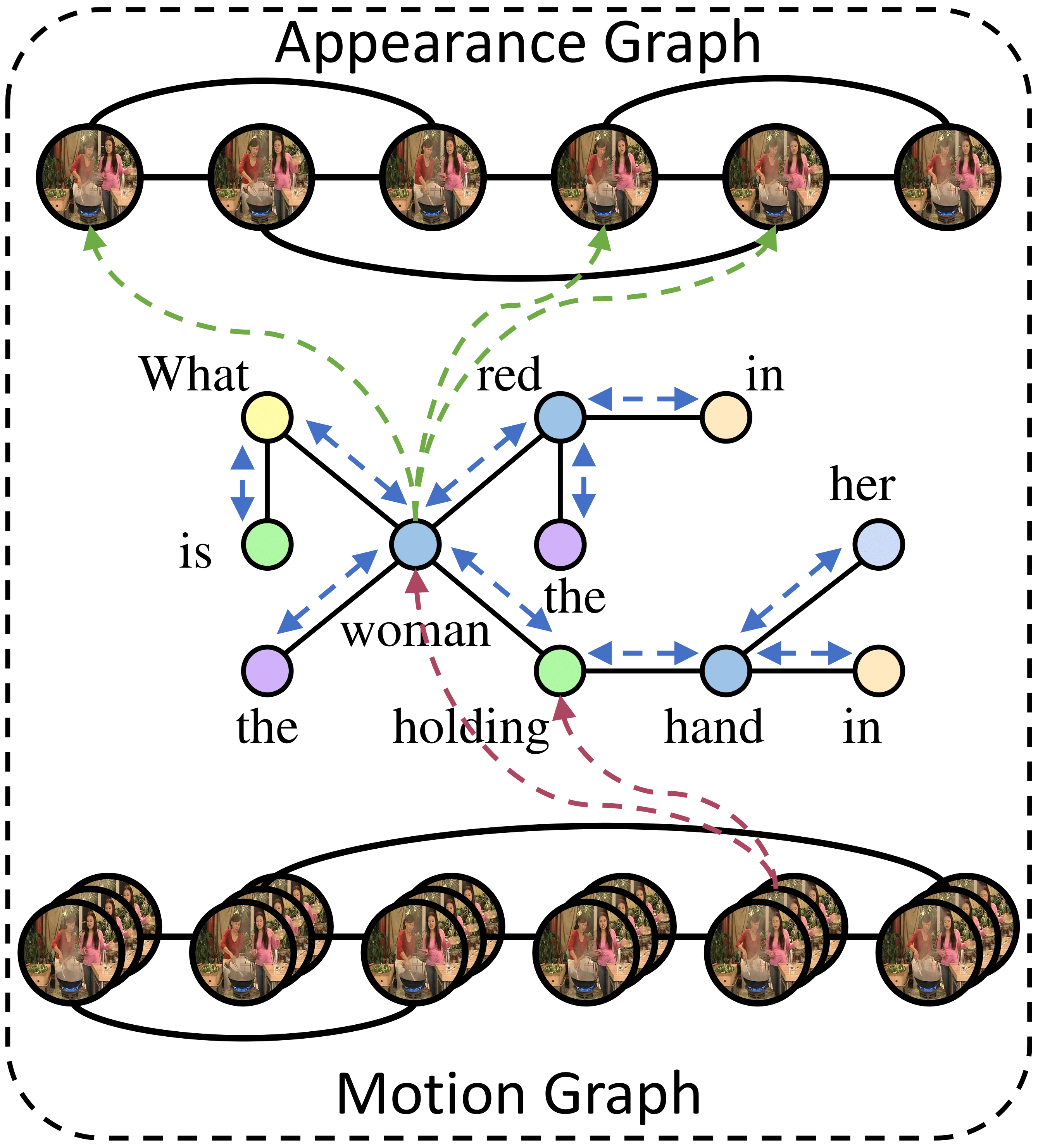}}\hfill
    \end{center}
       \caption{(a) An example for VideoQA. (b) Question-to-Visual (Q2V) interactions that each question node are propagated to visual nodes. (c) Visual-to-Visual (V2V) interactions that each visual node are associated with the relative visual nodes using the question bridge. Only motion-to-appearance (M2A) interaction is shown.}    \vspace{-10pt}
    \label{fig:1}
    \end{figure}
    
    Recent works~\cite{tgif-qa, ma18, comem, hra, hme, hcrn} have adopted the encoder-decoder structure.
    Typically, LSTM-based encoders~\cite{tgif-qa, comem, hra, hme} are used to encode the representations of video frames and a question into the visual and word sequence.
    The encoded representations are then incorporated to provide the answer with an attention mechanism.
    The several types of attention have shown promising results by learning the temporal relations between video frames~\cite{msvd-qa, hra}, the spatial relations between regions in every single frame~\cite{kim16, xiong16, ma18}, or spatiotemporal relations using appearance and motion representations~\cite{tgif-qa, comem}.
    Although these methods have suggested how to use the visual relationship for VideoQA, they still rely on learning positional relationships, not on in-depth semantic meaning, which makes capturing sophisticated appearance-motion or visual-question relations difficult.
    
    Meanwhile, methods to understand cross-modal relationships have been proposed for vision-language interaction tasks, such as image-text matching~\cite{li19iccv, gsmn} or video-text matching~\cite{chen20}, exploiting global~\cite{li17, nam17, wang19} or local~\cite{lee18, liu19} representations for visual and textual information.
    Similar approaches have also been adopted in VideoQA. 
    The global question representation has been used as a condition to learn a question-specific visual representation~\cite{zhao18, comem, hcrn}.
    For example, the global question feature vector was used to update the memory network to learn attention that attributed to the question in \cite{comem}.
    Le \etal~\cite{hcrn} proposed a hierarchical architecture that transforms a sequence of objects into a new array conditioned on the global question feature.
    The word-level features of the question have been treated as sequential data in the local approach~\cite{tgif-qa, msvd-qa, hme, psac, lgcn}.
    These approaches leveraged each word representation to learn visual attention~\cite{tgif-qa, msvd-qa, hra} or co-attention~\cite{ma18, hme, psac, lgcn} by fusing visual and word representations.
    However, the global approaches have learned coarse relations that frequently fail to capture video-word relations.
    The local approaches associate visual and word information based on co-occurrence statistics, not compositional semantics of the question.
    For instance, without semantic relations, the word ``\textit{woman}" of the question in \figref{fig:1}-(a) can incorrectly be correlated with all women in the video.
    On the other hand, compositional semantics clearly indicate from the phrase ``\textit{in the red}" that the question point to the left woman.
    
    To address these limitations, the consideration of grammatical dependencies between sentence words~\cite{marneffe06, marneffe08} has been raised.
    For visual question answering (VQA), Teney \etal~\cite{gsr} proposed structured representations that the input image and question are encoded as graphs to leverage compositional semantics of the question.
    For image-text matching, Liu \etal~\cite{gsmn} proposed a graph-structured network that construct graphs for the image and corresponding captions to find the fine-grained image-text correspondences using node-level and structure-level matching.
    Although the effectiveness of structured representations for image-text relations has been extensively demonstrated, it is still underexplored in VideoQA.
    
    In this paper, we propose a novel method, called \textit{Bridge to Answer}, that formulates structure-aware interaction for semantic relation modeling between crossmodal information, including appearance, motion, and question. 
    Contrary to existing approaches~\cite{comem, lgcn}, we construct not only appearance and motion graphs for video but also the question graph that represents compositional semantic relations between words.
    We perform question-to-video (Q2V) interactions that propagate the question node to its relevant visual nodes to learn question conditioned visual representations with visual-question relations, as shown in \figref{fig:1}-(b).
    Also, we apply visual-to-visual (V2V) interactions to visual graphs delivering each visual node to nodes in the relative visual graph to model appearance-motion relations.
    To utilize compositional semantic structure of the question, we use the question graph as an intermediate bridge, as shown in \figref{fig:1}-(c).
    We demonstrate the capability of the proposed method through extensive ablation studies and comparison with state-of-the-art methods on three datasets, including TGIF-QA~\cite{tgif-qa}, MSVD-QA~\cite{msvd-qa}, and MSRVTT-QA~\cite{msrvtt-qa}.


\section{Related Works}
\noindent\textbf{Video question answering (VideoQA)} has attracted intense attention over the past few years due to its applicability to human-robot interaction and video retrieval, etc.
    The existing methods have mostly been proposed to learn visual representations by leveraging video-question interactions.
    We summarize recent works according to the types of utilized interactions.
    Typically, the temporal attention has been learned by exploiting relationships between the appearance and question~\cite{tvqa, zhu17, li19}.
    Li~\etal~\cite{li19} proposed to learn co-attention between the appearance and question, and Li~\etal~\cite{psac} enhanced co-attention by using self-attention~\cite{vaswani17} mechanism.
    Some researchers have presented to capture fine-grained appearance-question interactions.
    Jin~\etal~\cite{jin19} introduced object-aware temporal attention that learns object-question interactions.
    Huang~etal~\cite{lgcn} also utilize frame and object features to enhance co-attention between the appearance and question.
    
    Since Jang~\etal~\cite{tgif-qa} proposed a two-stream architecture using appearance and motion features, researchers have focused on learning spatiotemporal attention that leverages motion, appearance, and question interactions.
    Developments of spatiotemporal attention have successfully been applied to various approaches including a multimodal fusion memory~\cite{hme}, co-memory attention~\cite{comem}, hierarchical attention~\cite{zhao17, zhao18}, multi-head attention~\cite{kim18}, and multi-step progressive attention~\cite{kim19, msvd-qa, song18}.
    The hierarchical structure that capture appearance-question and motion-question relations from the frame-level to segment-level have also been proposed by Zhao~\etal~\cite{zhao19} and Le~\etal~\cite{hcrn}.
    
    Although they have suggested methods that effectively learn relations between appearance and question or even motion, they still rely on positional relationships~\cite{psac}.
    Moreover, there have not been presented for capturing the relationship between appearance and motion with compositional semantics of the question.
    To address these limitations, we explicitly model appearance, motion, and the question as graphs.
    Our model learns question conditioned visual representations and mutually enhances appearance and motion representations by leveraging compositional semantics of the question.

    \begin{figure*}[t]
    \begin{center}
       \includegraphics[width=1\linewidth]{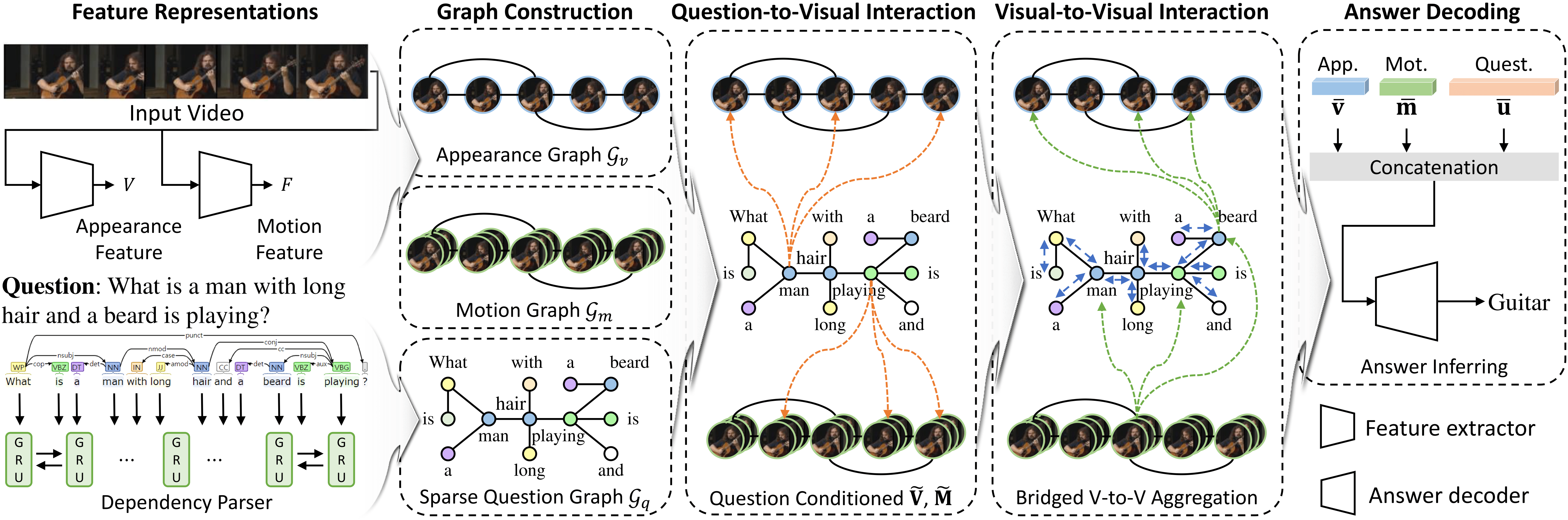}
    \end{center}
       \caption{The overall architecture of the proposed method for VideoQA. The visual and question representations are extracted to construct appearance, motion and question graphs. The graph nodes in each graph are propagated to nodes in another graph differentially through question-to-visual interactions and visual-to-visual interactions to learn question conditioned visual representation attributed to appearance and motion.} \vspace{-10pt}
    \label{fig:2}
    \end{figure*}
    
\noindent\textbf{Graph-structured vision-language interaction} has recently been studied to learn semantic relations between visual and textual information.
    Teney~\etal~\cite{gsr} firstly proposed to learn graph-structured representations of the input image and question for visual question answering (VQA).
    More recently, Li~\etal~\cite{li19iccv} proposed to learn relationships between regions in the input image using graph convolutional network (GCN)~\cite{gcn} and capture image-phrase correspondence for image-text matching.
    To enable fine-grained image-text matching, Liu~\etal\cite{gsmn} constructed a visual graph for the input image and textual graph with the compositional semantics of the caption, respectively.
    They successfully achieved state-of-the-art performance by learning correspondences between two structured graphs.
    Similarly, Chen~\etal~\cite{chen20} recently proposed a hierarchical graph reasoning method that learns fine-grained video-text correspondence.
    They composed hierarchical graphs for video and caption according to semantic roles, and performed global and local matching between two graphs.
    
    While these works have suggested methods that effectively learn visual-text relations with structured representations, they cannot be directly applied to VideoQA.
    To our knowledge, our work is the first attempt to perform relation reasoning between appearance and motion information of the video with compositional semantics of the question.

\section{Method}
    Given a video $\mathcal{V}$ and a question $q$, the VideoQA problem is generally formulated as follows:
    \begin{equation}
        \tilde{a} = \arg\max_{a \in \mathcal{A}}{\mathcal{F}_{\theta}(a | q, \mathcal{V})},
    \end{equation}
    where $\tilde{a}$ is the answer that can be inferred in answer space $\mathcal{A}$. $\theta$ is the set of model parameters of mapping function $\mathcal{F}$, which maps a pair of the video and question to the answer.
    We illustrate proposed method in \figref{fig:2}.
    We first extract feature representations from the video and the question, and construct graphs for appearance, motion, and question, respectively (\secref{sec:graph_construction}).
    The question nodes propagated to the visual graphs using question-to-visual interactions to learn question conditioned visual representations (\secref{sec:t-to-v_interaction}).
    Thereafter, the nodes in each visual graph are aggregated into relevant nodes in the relative visual graph over a question bridge to enhance visual representations by learning appearance-motion relations (\secref{sec:v-to-v_interaction}).
    Lastly, the final visual and question representations are concatenated and fed into the decoder to infer the answer (\secref{sec:answer_decoder}).
    \tabref{tab:notation} summarizes the notations used over our method.
    Following subsections, we depict the proposed method in details.
    \vspace{-10pt}
    \subsection{Feature Extraction and Graph Construction}\label{sec:graph_construction}
    \vspace{-10pt}
    \paragraph{Visual representations and visual graphs.}
    Similar to the previous works for videoQA~\cite{hcrn, lgcn}, we divide the video $\mathcal{V}$ of $L$ frames into $N$ uniform length clips $C = \lbrace C_1, ..., C_N \rbrace$, such that the length of each clip is $T = L/N$.
    To represent two types of information of the video, we extract frame-wise appearance feature vectors $\mathbf{V}$ and clip-wise motion feature vectors $\mathbf{M}$.
    In our work, $\mathbf{V}$ and $\mathbf{M}$ are extracted from the pretrained feature extractor (\eg, ResNet~\cite{resnet} and ResNeXt-101~\cite{resnext101}).
    The extracted features are fed into the linear feature transformation layers to project $\mathbf{V}$ and $\mathbf{M}$ into $d'$-dimensional feature space to obtain $\hat{\mathbf{V}} = \lbrace{\hat{\mathbf{v}}_{l} | \hat{\mathbf{v}}_{l} \in \mathbb{R}^{\text{d}'}}\rbrace_{l=1}^{L}$ and $\hat{\mathbf{M}} = \lbrace \hat{\mathbf{m}}_n \in \mathbb{R}^{\text{d}'} \rbrace_{n=1}^{N}$, respectively.
    
    With appearance and motion representations, we construct an appearance graph $\mathcal{G}_{v}$ and a motion graph $\mathcal{G}_{m}$ as undirected fully-connected graphs.
    The frames and clips are set to nodes, and each node is connected with all the other nodes in each graph with edges.
    The weight matrices $\mathbf{W}^{v}$ and $\mathbf{W}^{m}$, which represent node connections and their edge weights are computed by the affinity between node representations of $\hat{\mathbf{V}}$ and $\hat{\mathbf{M}}$ as
    \begin{equation}
        w_{ij}^{v} = \frac{\exp(\lambda \hat{\mathbf{v}}_{i}^{T} \hat{\mathbf{v}}_{j})}{\sum_{j = 0}^{L} \exp(\lambda \hat{\mathbf{v}}_{i}^{T} \hat{\mathbf{v}}_{j})}, \quad 
        w_{ij}^{m} = \frac{\exp(\lambda \hat{\mathbf{m}}_{i}^{T} \hat{\mathbf{m}}_{j})}{\sum_{j = 0}^{T} \exp(\lambda \hat{\mathbf{m}}_{i}^{T} \hat{\mathbf{m}}_{j})},
    \end{equation}
    where $w_{ij}^{v}$ and $w_{ij}^{m}$ indicate the edge weights between $i$-th and $j$-th node in each graph.
    $\lambda$ is a scaling factor.
    
    \begin{figure*}[t]
    \begin{center}
       \includegraphics[width=1\linewidth]{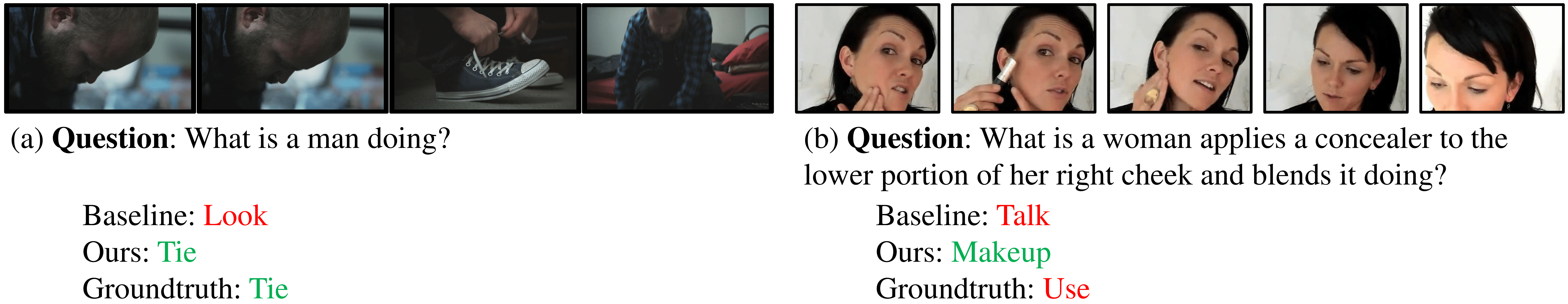}
    \end{center}
       \caption{Example questions for challenging conditions. (a) Sudden transitions of the scene lead to confusion in the model capturing the visual relation. (b) Long and complex question composition makes it difficult for the model to learn a properly conditioned visual representation. Our model with the capacity to capture relations of heterogeneous cross-modal graphs copes with these challenging cases.}    \vspace{-10pt}
    \label{fig:qual1}
    \end{figure*}
    
    \begin{table}[]
    \renewcommand\arraystretch{1.2}
    \begin{center}
        \begin{tabular}{|m{2.2cm} | m{5.1cm}|}
        \hline
        Notation & Role \\ \hline
          $\hat{\mathbf{V}}$, $\hat{\mathbf{M}}$       &   Input visual node representations   \\ \hline
             $\mathbf{U}$    &    Input question node representation  \\ \hline
             $\mathbf{W}^{v}$, $\mathbf{W}^{m}$, $\mathbf{W}^{q}$     &   Weight matrices of graphs   \\ \hline
            $\mathbf{S}^{v}$, $\mathbf{S}^{m}$     &   Q2V interaction matrix   \\ \hline
            $\tilde{\mathbf{V}}$, $\tilde{\mathbf{M}}$     &   Output of Q2V interaction   \\ \hline
             $\mathbf{U}_{b}^{v}$, $\mathbf{U}_{b}^{v}$    &    Bridged visual representations      \\ \hline
             $\hat{\mathbf{U}}_{b}^{v}$, $\hat{\mathbf{U}}_{b}^{m}$    &   Aggregated question representations   \\ \hline
            $\mathbf{S}_{b}^{v}$, $\mathbf{S}_{b}^{m}$     &   V2V interaction matrix   \\ \hline
            $\mathbf{V}^{f}$, $\mathbf{M}^{f}$     &   Output of V2V interaction   \\ \hline
        \end{tabular}
        \end{center}
        \caption{Notations of Bridge to Answer} \vspace{-10pt}
        \label{tab:notation}
    \end{table}

    \vspace{-15pt}
    \paragraph{Linguistic representations and question graph.}
    For the linguistic representation, we first embed all words in the question into $300$-dimensional vectors with pretrained word embeddings (\eg, GloVe~\cite{glove}).
    In the case of multiple choice questions, the words in answer candidates are also embedded.
    The embedded vectors are passed through a Bidirectional Gated Recurrent Unit (BiGRU) to establish the context dependency between words and are projected into the $d'$-dimensions feature space.
    The linguistic representations $\mathbf{U} = \lbrace \mathbf{u}_{i} | \mathbf{u}_{i} \in \mathbb{R}^{\text{d}'}\rbrace_{i=1}^{K}$ are obtained by concatenating the hidden states of forward and backward GRU at each time step, and applying a linear feature transformation, where $K$ is the number of words in a question.

    To take compositional semantic structure of the question into the question graph $\mathcal{G}_{q}$, we identify the semantic dependency within the question (and answer candidates) using Stanford CoreNLP~\cite{corenlp}.
    This parser is used to analyze the components in a sentence (\eg, nouns, verbs, or quantifiers) and parse their semantic dependencies (\eg, nominal subject or adjectival modifier).
    For example, given a question ``What is the woman in the red holding in her hand?", ``What", ``red", and ``holding" are semantically dependent with ``woman".
    Based on these dependencies, we set each word as a node and connect two nodes if they are semantically dependent.
    To obtain the weight matrix of the question graph, we compute the affinity matrix $\mathbf{E}$ of the question representation $\mathbf{U}$ as
    \begin{equation}
        e_{ij} = \frac{\exp(\lambda \hat{\mathbf{u}}_{i}^{T} \hat{\mathbf{u}}_{j})}{\sum_{j = 0}^{K} \exp(\lambda \hat{\mathbf{u}}_{i}^{T} \hat{\mathbf{u}}_{j})},
    \end{equation}
    where $e_{ij}$ indicates the affinity between the $i$-th and $j$-th question node and $\lambda$ is a scaling factor.
    Then the weight matrix $\mathbf{W}^q$ is represented by a Hadamard product between $\mathbf{E}$ and the adjacency matrix $\mathbf{A}$, followed by $L_2$ normalization, such that,
    \begin{equation}
        \mathbf{W}^q = ||\mathbf{E} \circ \mathbf{A}||_{2},
    \end{equation}
    where the adjacency matrix $\mathbf{A}$ represents the connectivity of the question graph.

\subsection{Question-to-Visual Interactions}\label{sec:t-to-v_interaction}
    The goal of question-to-visual (Q2V) interactions is to learn question conditioned visual representations by associating the question nodes with visual nodes and propagating question representations along visual edges through graph convolution layers~\cite{gcn}.
    The Q2V interactions are performed as question-to-appearance (Q2A) and question-to-motion (Q2M) interaction, respectively.
    Since Q2V interactions are symmetric operations on each graph except for the number of nodes, we describe Q2A interaction in detail and then roughly depict that on Q2M interaction.
    Specifically, we first obtain an interaction matrix $\mathbf{S}^{v}$ by applying softmax function to the affinity matrix between $\hat{\mathbf{V}}$ and $\mathbf{U}$ along the question axis, such that $\mathbf{S}^{v} = \text{softmax}_{\mathbf{U}}(\lambda \hat{\mathbf{V}}\mathbf{U}^{T})$.
    The interaction value $s_{ij}^{v}$ represents how much the $j$-th question node is associated with the $i$-th appearance node.
    All the question nodes are aggregated to the corresponding visual node with $\mathbf{S}^{v}$, followed by a fully connected (FC) layer, so that the aggregated appearance node representation is formulated as
    \begin{equation}
        \hat{\mathbf{v}}_{i}' = \sigma(\mathbf{W}_{f}^{v}(\hat{\mathbf{v}}_{i} + \sum_{j=1}^{K}{s_{ij}^{v}\mathbf{u}_{j}}) + b),
    \end{equation}
    where $\hat{\mathbf{v}}_{i}'$ is the $i$-th node representation of the aggregated appearance graph, $\mathbf{W}_{f}^{v}$ and $b$ are the learnable parameters of FC layer, and $\sigma(\cdot)$ is an activate function such as ReLU.
    
    Subsequently, we apply consecutive graph convolution layers that take $\hat{\mathbf{V}}'$ and the weight matrix $\mathbf{W}^{v}$ as inputs to propagate the aggregated node to its neighborhoods along the appearance edges.
    Formally, the output of Q2A interaction is represented as
    \begin{equation}
        \tilde{\mathbf{V}} = \mathcal{F}(\mathbf{W}^{v}, \hat{\mathbf{V}}' | \mathbf{W}_{g}^{v}),
    \end{equation}
    where $\mathbf{W}_{g}^{v}$ is the set of parameters of graph convolution layers.\footnote{We denote consecutive graph convolution layers as a feed-forward process $\mathcal{F}$.}
    
    Symmetrically, question-to-motion (Q2M) interaction, which is performed to obtain the question conditioned motion representation $\tilde{\mathbf{M}}$, can be formulated as
    \begin{equation}
        \begin{gathered}
            \hat{\mathbf{m}}_{i}' = \sigma(\mathbf{W}_{f}^{m}(\hat{\mathbf{m}}_{i} + \sum_{j=1}^{K}{s_{ij}^{m}\mathbf{u}_{j}}) + b),    \\
            \tilde{\mathbf{M}} = \mathcal{F}(\mathbf{W}^{m}, \hat{\mathbf{M}}' | \mathbf{W}_{g}^{m}),
        \end{gathered}
    \end{equation}
    where $\mathbf{S}^{m} = \text{softmax}_{\mathbf{U}}(\lambda \hat{\mathbf{M}}\mathbf{U}^{T})$ and $\mathbf{W}_{g}^{m}$ is the set of parameters of graph convolution layers.
    

\subsection{Visual-to-Visual Interactions}\label{sec:v-to-v_interaction}
    One of the most important capabilities for VideoQA is to capture and incorporate the relations between appearance and motion information.
    To realize this, we present visual-to-visual (V2V) interaction that learns semantic relationships between appearance and motion.
    Different from previous works~\cite{comem, hme} that appearance and motion information directly interact, we use the question graph as a bridge to leverage compositional semantics of the question.
    Since the structure of the question graph reflects semantic dependencies between words, the question conditioned visual node can effectively be delivered to the relative visual nodes along the question edges.
    
    The V2V interaction can be summarized as three-fold:
    1) one visual graph is bridged to the question graph, 
    2) the bridged node representation is propagated along the question edges through graph convolution layers and aggregated to the question graph, and 
    3) the aggregated question node is delivered to the relative visual graph.
    Concretely, motion-to-appearance (M2A) interaction begins by bridging motion and question graphs.
    A bridged motion representation, denoted as $\mathbf{U}_{b}^{m}$, can be obtained as a weighted combination of the question conditioned motion representation $\tilde{\mathbf{M}}$, where the weight is the interaction between $\mathbf{U}$ and $\tilde{\mathbf{M}}$, such that
    \begin{equation}
        \mathbf{U}_{b}^{m} = \text{softmax}_{\tilde{\mathbf{M}}}(\lambda \mathbf{U}\tilde{\mathbf{M}}^{T}) \tilde{\mathbf{M}}.
    \end{equation}
    The bridged representation is propagated to its neighbors along the question edges through graph convolution layers and aggregated to the question representation as
    \begin{equation}
        \hat{\mathbf{U}}_{b}^{m} = \mathbf{U} + \mathcal{F}(\mathbf{W}^{q}, \mathbf{U}_{b}^{m} | \mathbf{W}_{gb}^{m}),
    \end{equation}
    where $\mathbf{W}_{gb}^{m}$ is the set of trainable parameters of graph convolution layers.
    This form of the aggregated question graph enables the representation to have motion and question information simultaneously.
    
    Finally, the aggregated question node is delivered to the appearance graph to obtain a question conditioned appearance representation attributed to motion.
    The output of M2A interaction can be formulated by following equation: 
    \begin{equation}
        \begin{gathered}
            \mathbf{v}_{i}^{f} = \sigma(\mathbf{W}_{b}^{v}(\tilde{\mathbf{v}}_{i} + \sum_{j=1}^{K}{(s_{b}^{v})_{ij} (\hat{\mathbf{u}}_{b}^{m})_{j}}) + b), \\
            \mathbf{S}_{b}^{v} = \text{softmax}_{\hat{\mathbf{U}}_{b}^{m}}(\lambda \mathbf{\tilde{\mathbf{V}}}(\hat{\mathbf{U}}_{b}^{m})^{T}),
        \end{gathered}
    \end{equation}
    where $\mathbf{v}_{i}^{f}$ is the $i$-th node representation of the final appearance graph, $\mathbf{W}_{b}^{v}$ and $b$ are the parameters of FC layer.
    
    As a symmetric process, the node representation of the final motion graph $\mathbf{M}^{f}$ can be obtained with A2M interaction by following equations:
    \begin{equation}
        \begin{gathered}
            \mathbf{U}_{b}^{v} = \text{softmax}_{\tilde{\mathbf{V}}}(\lambda \mathbf{U} \tilde{\mathbf{V}}^{T}) \tilde{\mathbf{V}}, \\
            \hat{\mathbf{U}}_{b}^{v} = \mathbf{U} + \mathcal{F}(\mathbf{W}^{q}, \mathbf{U}_{b}^{v} | \mathbf{W}_{gb}^{v}),   \\
            \mathbf{m}_{i}^{f} = \sigma(\mathbf{W}_{b}^{m}(\tilde{\mathbf{m}}_{i} + \sum_{j=1}^{K}{(s_{b}^{m})_{ij} (\hat{\mathbf{u}}_{b}^{v})_{j}}) + b),  \\
            \mathbf{S}_{b}^{m} = \text{softmax}_{\hat{\mathbf{U}}_{b}^{v}}(\lambda \mathbf{\tilde{\mathbf{M}}}(\hat{\mathbf{U}}_{b}^{v})^{T}),
        \end{gathered}
    \end{equation}
    where $\mathbf{W}_{gb}^{v}$ and $\mathbf{W}_{b}^{m}$ are the weight parameters of graph convolution layers and FC layer, respectively.
      
    We apply an average pooling to the final visual node representations along the temporal axis to vectorize the representations, and concatenate them to make the incorporated visual representation $\mathbf{o}$:
    \begin{equation}
        \begin{gathered}
            \bar{\mathbf{v}} = \frac{1}{L} \sum_{l = 1}^{L} {\mathbf{v}_{l}^{f}}, \quad \bar{\mathbf{m}} = \frac{1}{N} \sum_{n = 1}^{N} {\mathbf{m}_{n}^{f}},    \\
             \mathbf{o} = [\bar{\mathbf{v}}; \bar{\mathbf{m}}],
        \end{gathered}
    \end{equation}
    where $[\cdot;\cdot]$ denotes concatenation operation. 
    
    \begin{table}
    \begin{center}
    \begin{tabular}{
    >{\raggedright}m{0.22\linewidth}>{\centering}m{0.13\linewidth}
    >{\centering}m{0.13\linewidth}>{\centering}m{0.13\linewidth}>{\centering}m{0.13\linewidth}}
        \hlinewd{0.8pt}
        Model & Action & Trans. & Frame & Count  \tabularnewline
        \hline \hline
        ST-TP~\cite{tgif-qa} &  62.9    &   69.4    &   49.5    &   4.32   \tabularnewline
        Co-mem~\cite{comem} &   68.2    &   74.3    &   51.5    &   4.10   \tabularnewline
        PSAC~\cite{psac}    &   70.4    &   76.9    &   55.7    &   4.27    \tabularnewline
        HME~\cite{hme}      &   73.9    &   77.8    &   53.8    &   4.02   \tabularnewline
        L-GCN~\cite{lgcn}   &   74.3    &   81.1    &   56.3    &   3.95  \tabularnewline
        QueST~\cite{quest}   &   75.9    &   81.0    &   \textbf{59.7}    &   4.19  \tabularnewline
        HCRN~\cite{hcrn}    &   75.0    &   81.4    &   55.9    &   3.82     \tabularnewline    \hline
        \textbf{Ours}       &   \textbf{75.9}   &   \textbf{82.6}   &   57.5    &  \textbf{3.71}    \tabularnewline
        \hlinewd{0.8pt}
        \end{tabular}
    \end{center}
    \vspace{-5pt}
    \caption{Performance comparison for several tasks on TGIF-QA~\cite{tgif-qa} dataset with state-of-the-art methods.
    The lower the better for count.
    }\label{tab:tgif-qa}\vspace{-10pt}
    \end{table}
    
\subsection{Answer Decoder and Loss Functions}\label{sec:answer_decoder}
    Following previous works~\cite{hme, hcrn}, we use different answer decoders depending on the type of question.
    Specifically, we treat open-ended questions as a multi-label classification problem.
    The decoder takes the final visual representation $\mathbf{o}$ and the averaged question representation $\bar{\mathbf{u}}$ to compute label probabilities $p \in \mathbb{R}^{|\mathcal{A}|}$:
    \begin{equation}\label{eq:y}
        \begin{gathered}
            y = \sigma(\mathbf{W}_{2}[\mathbf{o}, \mathbf{W}_{1} \bar{\mathbf{u}} + b] + b), \\
            y' = \sigma(\mathbf{W}_{y}y + b),   \\
            p = \text{softmax}(\mathbf{W}_{y'}y' + b),
        \end{gathered}
    \end{equation}
    where $\mathbf{W}_{1}$, $\mathbf{W}_{2}$, $\mathbf{W}_{y}$, and $\mathbf{W}_{y'}$ are of learnable parameters of the decoder.
    We employ the cross-entropy loss for open-ended questions.

    We treat repetition count task as a linear regression problem that the decoder takes $y'$ in ~\equref{eq:y} as an input and applies a rounding function for integer count results.
    The Mean Squared Error (MSE) is employed as the loss function.
    
    For multiple choice question types (\ie, repeating action and state transition), $|\mathcal{A}|$ answer candidates are used to make a set of visual representations corresponding to each candidate in the same way with the question.
    As a result, we have a set of final visual representations, $\mathbf{o}$ conditioned by the question, and $\lbrace \mathbf{o}_{i}^{a} \rbrace_{i=1}^{|\mathcal{A}|}$ conditioned by answer candidates.
    The classifier for multiple choice question takes $\mathbf{o}$, $\mathbf{o}_{i}^{a}$, $\bar{\mathbf{u}}$, and answer candidates $\bar{\mathbf{a}}_{i}$ to output probabilities for candidates as follows:
    \begin{equation}
        \begin{gathered}
        y_{i} = [\mathbf{o}, \mathbf{o}_{i}^{a}, \mathbf{W}_{w}\bar{\mathbf{u}} + b, \mathbf{W}_{a}\bar{\mathbf{a}}_{i} + b], \\
        y'_{i} = \sigma(\mathbf{W}_{y}y_{i} + b),   \\
        s_{i} = \mathbf{W}_{y'}y'_{i} + b,
        \end{gathered}
    \end{equation}
    where $\mathbf{W}_{1}$, $\mathbf{W}_{a}$, $\mathbf{W}_{y}$, and $\mathbf{W}_{y'}$ are of learnable parameters the decoder.
    Then, the candidate with the largest $s$ value is selected as the answer such that,
    \begin{equation}
        \tilde{a} = \arg\max_{i} s_{i}.
    \end{equation}
    We employ the hinge loss~\cite{hingeloss} between incorrect answer score $s^{n}$ and correct answer score $s^{p}$, $\max(0, 1 + s^{n} - s^{p})$, as the loss function.

    \begin{table}
    \begin{center}
    \begin{tabular}{
    >{\raggedright}m{0.25\linewidth}>{\centering}m{0.23\linewidth}
    >{\centering}m{0.23\linewidth}}
        \hlinewd{0.8pt}
        Model & MSVD-QA & MSRVTT \tabularnewline
        \hline \hline
        AMU~\cite{msvd-qa} &  32.0    &   32.5    \tabularnewline
        HRA~\cite{hra}  &   34.4    &   35.0    \tabularnewline
        Co-mem~\cite{comem} &   31.7    &   31.9    \tabularnewline
        HME~\cite{hme}      &   33.7    &   33.0    \tabularnewline
        L-GCN~\cite{lgcn}   &   34.3    &   -    \tabularnewline
        QueST~\cite{quest}    &   36.1    &   34.6    \tabularnewline
        HCRN~\cite{hcrn}    &   36.1    &   35.6    \tabularnewline \hline
        \textbf{Ours}       &   \textbf{37.2}   &   \textbf{36.9}   \tabularnewline
        \hlinewd{0.8pt}
        \end{tabular}
    \end{center}
    \caption{Performance comparison for open-ended questions on MSVD-QA~\cite{msvd-qa} and MSRVTT-QA~\cite{msrvtt-qa} datasets with state-of-the-art methods.
    }\label{tab:msvd-qa}\vspace{-10pt}
    \end{table}
 
\section{Experiment}
\subsection{Datasets}

    \paragraph{TGIF-QA~\cite{tgif-qa}} dataset contains $72K$ animated GIF files and $165K$ question answer pairs.
    The dataset provides four kinds of tasks that address the unique properties of videos.
    \textit{Repetition Count} is to retrieve number of occurrences of an action.
    \textit{Repeating action} is a task to identify an action repeated for a given number of times among multiple choices.
    \textit{State Transition} is a multiple choice task to identify an action regarding the temporal order of action state.
    \textit{Frame QA} is to find a particular frame in a video that can answer the questions.
    \vspace{-15pt}
    \paragraph{MSVD-QA~\cite{msvd-qa}} dataset contains $1,970$ short clips and $50,505$ question answer pairs.
    The questions are composed of five types, including what, who, how, when, and where.
    \vspace{-15pt}
    \paragraph{MSRVTT-QA~\cite{msrvtt-qa}} dataset contains $10K$ videos and $243K$ question answer pairs.
    While types of questions are the same with MSVD-QA dataset, the contents of the videos in MSRVTT-QA are more complex and the lengths of the videos are much longer from 10 to 30 seconds.
    
    For the evaluation metrics, we use Mean Squared Error (MSE) for repetition count on TGIF-QA dataset and use accuracy for all the other experiments.

\subsection{Implementation Details}
    We divide the video into 8 clips containing 16 frames in each clip by default.
    Following the previous work~\cite{hcrn}, the long videos in MSRVTT-QA are additionally divided into 24 clips.
    We train our model for 25 epochs with a batch size of 16 for TGIF-QA and MSVD-QA datasets, and of 4 for MSRVTT-QA dataset.
    The learning rate is set to $10^{-4}$ and decayed by half for every 5 epochs.
    The reported results are at the epoch showing the best validation accuracy.

    \begin{table}[]
    \begin{center}
    \begin{tabular}{lcccc}
    \hlinewd{0.8pt}
    Model                               & Act.                 &        Trans.               &      F.QA                 &            Count                \\ \hline\hline
    \textbf{Input conditioning} &                      &                      &                      &                      \\
    \quad w/o appearance              &         72.8             &          77.2            &         -             &          4.01            \\
    \quad w/o motion                  &            68.2          &           75.5           &          57.3            &          4.21            \\ \hline
    \textbf{Interaction}        &                      &                      &                      &                      \\
    \quad w/o Q2V, V2V                &          69.4            &         75.3             &          51.4            &         3.97             \\
    \quad w/o Q2A                     &          73.9            &          80.1            &           52.7           &          3.87            \\
    \quad w/o Q2M                     &         73.1             &            78.2          &           56.8           &            3.90          \\
    \quad w/o Q2V                     &         72.3             &          76.3            &           52.6          &             3.95         \\
    \quad w/o A2M                     &           74.7           &          81.4            &           56.1           &            3.84          \\
    \quad w/o M2A                     &         74.8             &          80.9            &           56.9           &            3.86          \\
    \quad  w/o V2V                     &            74.1          &         78.5             &          56.5            &           3.89           \\ \hline
    \textbf{Bridge conditioning}        &                      &                      &                      &                      \\
    \quad w/o bridge                &          75.1            &         81.7             &          56.9            &          3.83            \\    \hline
    \textbf{Parameter $\lambda$}  &                       &                       &                       &                           \\
    \quad $\lambda = 1$               &       75.1              &       81.5                &           56.2            &           3.80            \\
    \quad $\lambda = 5$               &       75.3              &       81.8                &           57.2            &           3.77            \\
    \quad $\lambda = 10$               &      75.9               &      82.6                 &          57.5             &          3.71             \\
    \quad $\lambda = 20$               &        75.4             &          82.2             &            57.1           &              3.73         \\  \hline
    \quad Full model                  &         75.9            &             82.6           &            57.5           &            3.71  \\ \hlinewd{0.8pt}
    \end{tabular}
    \end{center}
    \caption{Ablation studies for input conditioning, interaction, and the value of $\lambda$ on TGIF-QA dataset~\cite{tgif-qa}. Act.: Action; Trans.: Transition; F.QA: Frame QA. When not explicitly specified, we use $\lambda = 10$. The lower the better for count.}\label{tab:ablation_tgif}\vspace{-10pt}
    \end{table}

\subsection{Experimental Results}

    We compare our model with several state-of-the-art methods on TGIF-QA, MSVD-QA, and MSRVTT-QA datasets.\footnote{Reported values of the other methods are taken from the original papers and \cite{hcrn}}
    For TGIF-QA dataset, we compare our model with \cite{tgif-qa, comem, psac, hme, lgcn, quest, hcrn} in \tabref{tab:tgif-qa}.
    We display evaluation results over four tasks, including repeating action, state transition, frameQA, and repetition count.
    The results show that our model achieves state-of-the-art performance and outperforms the existing methods on all tasks except for FrameQA task.
    
    For MSVD-QA and MSRVTT-QA datasets, our model is compared with \cite{msvd-qa, hra, comem, hme, lgcn, quest, hcrn} in \tabref{tab:msvd-qa}.
    Since these datasets provide open-ended questions, they are referred to as highly challenging benchmarks compared to the TGIF-QA dataset.
    Our model achieves $37.2\%$ and $36.9\%$ accuracy, outperforming the existing approaches by $1.1\%$ and $1.3\%$ for accuracy, respectively.
    
    We provide qualitative results for two challenging examples in \figref{fig:qual1}.
    The first example shows that a sudden transition of the scene causes the problem of capturing semantic relations.
    Our model handles this problem by learning in-depth semantic relations, not positional relations.
    The second example reflects the case in which a long and complex question has given\footnote{Groundtruth is probably incorrectly annotated.}.
    Our model successfully analyzes this long and complex question by explicitly modeling the compositional semantics of the question.

    \begin{figure*}[t]
    \begin{center}
       \includegraphics[width=1\linewidth]{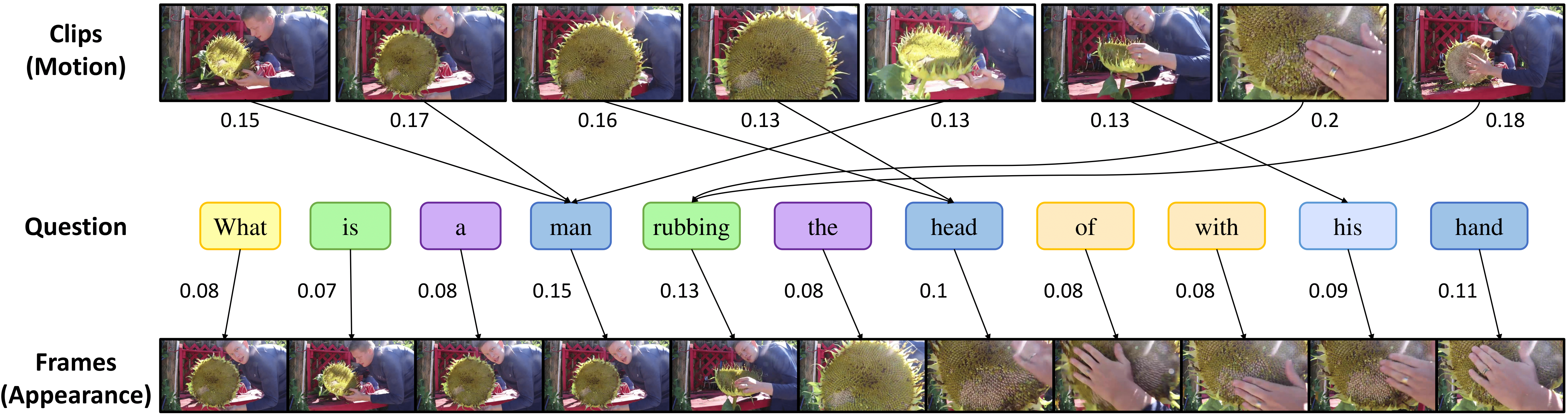}
    \end{center}
       \caption{Visualization of M2A interaction. Although any two graphs are fully connected by interaction value, we only indicate the connection with the largest value in each interaction matrix for visibility. Note that the clips are represented by frames sampled from each clip.}
    \label{fig:qual2}\vspace{-10pt}
    \end{figure*}

    \begin{table}
    \begin{center}
    \begin{tabular}{
    >{\raggedright}m{0.25\linewidth}>{\centering}m{0.23\linewidth}
    >{\centering}m{0.23\linewidth}}
        \hlinewd{0.8pt}
        Parameter $\lambda$ & MSVD-QA & MSRVTT \tabularnewline
        \hline \hline
        $\lambda = 1$ &  35.3    &   33.8    \tabularnewline
        $\lambda = 5$  &   36.9    &   35.0    \tabularnewline
        $\lambda = 10$ &   $\mathbf{37.2}$    &   36.6    \tabularnewline
        $\lambda = 20$      &   36.5    &   $\mathbf{36.9}$    \tabularnewline
        \hlinewd{0.8pt}
        \end{tabular}
    \end{center}
    \caption{Performance comparison with different $\lambda$ values on MSVD-QA~\cite{msvd-qa} and MSRVTT-QA~\cite{msrvtt-qa} datasets.
    }\label{tab:lambda_msvd}\vspace{-10pt}
    \end{table}

\subsection{Ablation Studies}
    To validate the effectiveness of components within our model, we conduct extensive ablation studies on TGIF-QA~\cite{tgif-qa}, as shown in \tabref{tab:ablation_tgif}.
    Ablation studies widely cover the results according to input conditioning, interactions, question bridge, and the value of the parameter $\lambda$.
    The overall result verify that all components affect performance, and even any direction of interaction contribute to the performance improvement.
    The detailed analyzes are described below.
    \vspace{-10pt}
    \paragraph{Effect of input conditioning.}
    We study the effect of the input condition with following settings:\\
    $\blacktriangleright\textit{w/o appearance}$: Remove appearance feature from full model. Q2A and V2V interaction also been removed. Since the frames are not used in this setting, we do not measure the performance for FrameQA.
    $\blacktriangleright\textit{w/o motion}$: Remove motion feature from full model. Q2M and V2V interaction also been removed.
    We find that while the absence of either appearance or motion feature is critical to the performance, motion contributes more to the performance in action-related tasks.
    The results of FrameQA show that motion information does not play an important role in tasks where appearance information of the frame is important.
    \vspace{-15pt}
    \paragraph{Effect of interactions.}
    To investigate the effectiveness of each interaction (\eg, Q2A or A2M), we evaluate our model with all possible combination of interactions:\\
    $\blacktriangleright\textit{w/o Q2V}$: Not using question conditioned visual representations and performing V2V interaction only with the question bridge.
    $\blacktriangleright\textit{w/o Q2A}$ or $\textit{w/o Q2M}$: Using only one question conditioned visual representation (\ie, using $\hat{\mathbf{V}}$ and $\tilde{\mathbf{M}}$, or $\tilde{\mathbf{V}}$ and $\hat{\mathbf{M}}$). $\blacktriangleright\textit{w/o V2V}$: No interaction between the two visual representations.
    $\blacktriangleright\textit{w/o A2M}$ or $\textit{w/o M2A}$: Using one-way V2V interaction (\ie, motion to appearance or appearance to motion only).
    
    Overall we find that the absence of any direction of interaction significantly degrades the performance on all tasks.
    Specifically, Q2V works as a primary component for VideoQA.
    This is not surprising given that learning question conditioned visual representations is one of the most important ingredient for VideoQA.
    The notable performance degradation due to the ablation of Q2M is shown in all the tasks except for FrameQA.
    
    We also find that A2M and M2A are complemented each other from the results that promising performance can be achieved with only one-way interaction.
    To analyze V2V interaction, we display the visualization for the connectivity of motion-question and question-appearance as shown in \figref{fig:qual2}.
    We depict M2A interaction only and represent the clips as frames sampled at each clip for visibility.
    The clips and words are placed according to temporal and word order, respectively, and corresponding frames of each word are placed regardless of temporal order.
    The connections indicate that two nodes are associated with the maximum interaction value.
    For example, the first clip is associated with the word ``man" by the interaction value of 0.15, and the word ``man" is related to the fourth frame by the interaction value of 0.15.
    Note that, when all the nodes are connected with uniformly distributed weights, the motion-question interaction value and the question-appearance interaction value is 0.09 and 0.008, respectively.
    The results show that the relavant nodes are connected with relatively high interaction values, and the question node is also connected with the appearance node through semantic relation.
    \vspace{-10pt}
    \paragraph{Effect of the question bridge.}
    The question bridge is a key component to leverage the compositional semantics of the question.
    To verify the effectiveness, we conduct ablation study for the question bridge.
    The V2V interactions without the question bridge are performed by directly aggregating the relative node representations based on the affinity value between appearance and motion nodes.
    For instance, the output of the M2A interaction without the bridge is obtained by
    \begin{equation}
        \mathbf{v}^{f}_{i} = \sigma(\mathbf{W}_{wob}^{v}(\tilde{\mathbf{v}}_{i} + \sum_{j = 1}^{T} {(s_{wob}^{v})_{ij}\tilde{\mathbf{m}}_{j}}) + b),   
    \end{equation}
    where $\mathbf{S}_{wob}^{v} = \text{softmax}_{\tilde{\mathbf{M}}}(\lambda \tilde{\mathbf{V}} \tilde{\mathbf{M}}^{T}).$

    The results at each task demonstrate the advantage of the bridged architecture with the performance improvement of $0.5\%$, $0.9\%$, $0.6\%$ for accuracy, and $0.12$ for MSE value, respectively.
    \vspace{-10pt}
    \paragraph{Effect of $\mathbf{\lambda}$.}
    The scaling parameter $\lambda$ adjusts the relative weight of different nodes in Q2V and V2V interactions and the edge weight of graphs.
    The large value of $\lambda$ distills nodes highly correlated to the specific node and filters out irrelevant nodes.
    Contrary to this, the small value of $\lambda$ is difficult to distinguish relevant nodes.
    Therefore, it is important to properly set the value of $\lambda$. 
    To investigate the performance with various $\lambda$ values, we measure VideoQA performance by setting the $\lambda$ as 1, 5, 10, 20.
    Not surprisingly, Lager $\lambda$ shows better performance compared to $\lambda = 1$.
    
    We additionally evaluate our model according to $\lambda$ values on MSVD-QA~\cite{msvd-qa} and MSRVTT-QA~\cite{msrvtt-qa} datasets.
    As shown in \tabref{tab:lambda_msvd}, our model yield highest performance when $\lambda = 10$ on MSVD-QA.
    The results on MSRVTT-QA show that $\lambda = 20$ brings out the best performance.
    The different optimal value of $\lambda$ on two datasets might be caused by different lengths of videos.
    
\section{Conclusion}
    We proposed a novel method for VideoQA, called Bridge to Answer, that constructs heterogeneous multimodal graphs and learns relations between visual and question graphs to learn question conditioned visual representations attributed to appearance and motion.
    In the process, in-depth semantic relations between visual and question graphs are encapsulated to visual representations using question-visual interactions.
    The relations between appearance and motion graphs are modulated by compositional semantics of the question as a bridge to effectively enhance each relative visual representation.
    This bridged structure allows a model robust to the scene composition and sophisticated structure of the question.
    Our model was evaluated on several VideoQA benchmarks, including TGIF-QA, MSVD-QA, and MSRVTT-QA, achieving state-of-the-art performance.
    
\section*{Acknowledgement}
    This work was supported by Institute of Information communications Technology Planning \& Evaluation (IITP) grant funded by the Korea government(MSIT) (No.2020-0-00056, To create AI systems that act appropriately and effectively in novel situations that occur in open worlds)










{\small
\bibliographystyle{ieee_fullname}
\bibliography{egbib}
}

\end{document}


	\title{Bridge to Answer: Structure-aware Graph Interaction Network \\for Video Question Answering \\ -Supplementary Materials-}

	\author{Jungin Park \quad\quad\quad Jiyoung Lee \quad\quad\quad Kwanghoo Sohn\thanks{Corresponding author.}  \\
		Yonsei University, Seoul, South Korea   \\
		{\tt\small $\lbrace$ newrun, easy00, khsohn $\rbrace $@yonsei.ac.kr}
	}

	\maketitle

    In this document, we briefly summarize graph convolution networks~\cite{gcn}, and provide more details of \textit{Bridge to Answer} and more qualitatively results that demonstrate the advantage of our method.
    
\section{Graph Convolution Networks}
    A graph $\mathcal{G}$ can be represented by a tuple $\mathcal{G} = \{\mathcal{V}, \mathcal{E}\}$, where $\mathcal{V}$ is the set of vertices and $\mathcal{E}$ is the set of edges representing the connectivity between vertices, such that the vertices $v_i$ and $v_j$ are connected with the edge weight $e_{ij}$.
    An adjacency matrix $\mathbf{A}$ contains the connectivity between vertices and their edge weights in a fully-connected or a sparse matrix form.
    The graph convolution networks (GCNs)~\cite{gcn} have been proposed to learn richer representations of vertices by aggregating the representations from their neighborhoods.
    
    In standard GCNs, the node representations $\mathbf{X}$ and the adjacency matrix $\mathbf{A}$ are taken as inputs of graph convolution operation.
    The output of the graph convolution operation is then obtained by following equation:
    \begin{equation}
        \mathbf{Z} = \sigma(\mathbf{D}^{-1/2}\hat{\mathbf{A}}\mathbf{D}^{-1/2}\mathbf{X}\mathbf{W}),
    \end{equation}
    where $\mathbf{Z}$ is the output node representations, $\sigma(\cdot)$ is an activation function such as ReLU, and $\hat{\mathbf{A}}$ is the summation of $\mathbf{A}$ and the identity matrix $\mathbf{I}$, such that $\hat{\mathbf{A}} = \mathbf{A} + \mathbf{I}$.
    $\mathbf{D}$ is a diagonal matrix of $\hat{\mathbf{A}}$ and $\mathbf{W}$ is a trainable weight matrix of the graph convolution layer, respectively.
    
    In this paper, we construct fully-connected appearance and motion graphs and a sparse question graph and apply consecutive graph convolution layers to perform question-to-visual and visual-to-visual interactions.

\section{Channel Configurations}

\begin{table}[t!]
    \begin{center}
    \begin{tabular}{
			>{\raggedright}m{0.3\linewidth} |
			>{\raggedright}m{0.5\linewidth}}
         \hlinewd{0.8pt}
         Notation   &   Channel \tabularnewline
         \hline\hline
         $\hat{\mathbf{V}}$, $\hat{\mathbf{M}}$ &   $\mathbb{R}^{8 \times 16 \times 512}$, $\mathbb{R}^{8 \times 512}$    \tabularnewline\hline
         $\mathbf{U}$   &    $\mathbb{R}^{w \times 512}$   \tabularnewline\hline
         $\mathbf{W}^{v}$, $\mathbf{W}^{m}$, $\mathbf{W}^{m}$   &    $\mathbb{R}^{128 \times 128}$, $\mathbb{R}^{8 \times 8}$, $\mathbb{R}^{w \times w}$   \tabularnewline\hline
         $\mathbf{S}^{v}$, $\mathbf{S}^{m}$   &    $\mathbb{R}^{128 \times w}$, $\mathbb{R}^{8 \times w}$   \tabularnewline\hline
         $\tilde{\mathbf{V}}$, $\tilde{\mathbf{M}}$ &   $\mathbb{R}^{8 \times 16 \times 512}$, $\mathbb{R}^{8 \times 512}$    \tabularnewline\hline
         $\mathbf{U}_{b}^{v}$, $\mathbf{M}_{b}^{m}$ &   $\mathbb{R}^{w \times 512}$, $\mathbb{R}^{w \times 512}$    \tabularnewline\hline
         $\hat{\mathbf{U}}_{b}^{v}$, $\hat{\mathbf{M}}_{b}^{m}$ &   $\mathbb{R}^{w \times 512}$, $\mathbb{R}^{w \times 512}$    \tabularnewline\hline
         $\mathbf{S}_{b}^{v}$, $\mathbf{S}_{b}^{m}$ &   $\mathbb{R}^{8 \times 16 \times w}$, $\mathbb{R}^{8 \times w}$    \tabularnewline\hline
         $\mathbf{V}^{f}$, $\mathbf{M}^{f}$ &   $\mathbb{R}^{8 \times 16 \times 256}$, $\mathbb{R}^{8 \times 256}$    \tabularnewline\hline
         $\bar{\mathbf{v}}$, $\bar{\mathbf{m}}$ &   $\mathbb{R}^{256}$, $\mathbb{R}^{256}$    \tabularnewline\hline
    \end{tabular}
    \end{center}
    \caption{Notations and channel configuration}
    \label{tab:configuration}
\end{table}

 \begin{table}[t!]
        \begin{center}
        \begin{tabular}{
                >{\raggedright}m{0.17\linewidth} |
    			>{\raggedright}m{0.22\linewidth} |
    			>{\raggedright}m{0.4\linewidth}}
             \hlinewd{0.8pt}
             Equ. & Notation   &   Dimensions \tabularnewline
             \hline\hline
            (5), (7) & $\mathbf{W}_{f}^{v}$, $\mathbf{W}_{g}^{v}$  &   $\mathbb{R}^{512 \times 512}$    \tabularnewline\hline
            (6), (7) & $\mathbf{W}_{f}^{m}$, $\mathbf{W}_{g}^{m}$  &    $\lbrace \mathbb{R}^{512\times512}, \mathbb{R}^{512 \times 256}\rbrace$ \tabularnewline \hline
             (9), (11) & $\mathbf{W}_{gb}^{v}$, $\mathbf{W}_{gb}^{m}$   &  $\lbrace \mathbb{R}^{512\times512}, \mathbb{R}^{512 \times 256}\rbrace$   \tabularnewline\hline
             (10), (11) & $\mathbf{W}_{b}^{v}$, $\mathbf{W}_{b}^{m}$ & $\mathbb{R}^{512 \times 512}$  \tabularnewline \hline
             (13) & $\mathbf{W}_{1}$, $\mathbf{W}_{2}$   &    $\mathbb{R}^{512 \times 512}$, $\mathbb{R}^{1024 \times 512}$   \tabularnewline\hline
             (13) & $\mathbf{W}_{y}$, $\mathbf{W}_{y'}$   &    $\mathbb{R}^{512 \times 256}$, $\mathbb{R}^{256 \times W}$   \tabularnewline\hline
             (14) & $\mathbf{W}_{w}$, $\mathbf{W}_{a}$   &    $\mathbb{R}^{512 \times 512}$   \tabularnewline\hline
             (14) & $\mathbf{W}_{y}$, $\mathbf{W}_{y'}$   &    $\mathbb{R}^{2048 \times 512}$, $\mathbb{R}^{512 \times 1}$   \tabularnewline
             \hlinewd{0.8pt}
             
        \end{tabular}
        \end{center}
        \caption{Dimensional configuration for the weights}
        \label{tab:weight_config}
    \end{table}

    Our method is divided into three parts: \textit{1) Graph construction} to generate appearance, motion, and question graphs,
    \textit{2) question-to-visual interactions} to make question conditioned visual representations (\ie, question-to-appearance and question-to-motion), and
    \textit{3) visual-to-visual interactions} to propagate each visual information to relative visual graph (\ie, appearance-to-motion and motion-to-appearance).
    To clarify the usage of each component in our method, we provide the overall notations and channel configurations as shown in \tabref{tab:configuration}.
    
    In addition, we describe the dimensional configuration for the weights used in Q2V and V2V interactions, as shown in \tabref{tab:weight_config}. 


    \begin{figure*}[t]
        \begin{center}
           \includegraphics[width=1\linewidth]{figure/supp_qual.pdf}
        \end{center}
           \caption{Qualitative comparisons with the state-of-the-art method~\cite{hcrn}. The results show the advantages of our method in various challenging cases: (a) Requiring more accurate answer. (b) Inferring action by appearance, not motion. (c) Capturing the rapid transition of the object. (d) Capturing the slow transition of the object. (e) Finding a target in multiple objects. (f) Processing a long and complex question.}
        \label{fig:supp_qual}
    \end{figure*}

\section{More Results}
    In this supplementary, we provide the more quantitative comparisons with the state-of-the-art method~\cite{hcrn}. As shown in \figref{fig:supp_qual}, our model shows the advantages in various challenging cases.
    When the answer candidates are ambiguous, our model finds a more plausible answer.
    For example, the candidates ``hit an object" and ``slap tail" in \figref{fig:supp_qual}-(a) both can be correct answers.
    Our model infer a more plausible answer ``slap tail".
    \figref{fig:supp_qual}-(b) shows an example of \textit{inferring action by appearance}.
    The action of ``turn off lights" is more related to appearance information, not motion.
    Since we learn the question conditioned appearance representation attributed to motion, our model adequately captures the action with the appearance changes.
    In addition, our model captures both rapid and slow transitions of an object as shown in \figref{fig:supp_qual}-(c) and (d).
    The examples of the last row in \figref{fig:supp_qual} show the advantages of using the question graph.
    By associating question and visual graphs and propagating information, our model capture the object referred to in the question when multiple objects are in the video as shown in \figref{fig:supp_qual}-(e).
    Also, more accurate processing for a long and complex question is possible by constructing the question graph that considers the compositional semantics of the question.
    
    \begin{figure*}[t]
        \begin{center}
          \includegraphics[width=1\linewidth]{rebuttal/rebuttal_qual.pdf}
        \end{center}
          \caption{Visualization of A2M interaction. Although any two graphs are fully connected by interaction value, we only indicate the connection with the largest value in each interaction matrix for visibility. Note that the clips are represented by frames sampled from each clip.}
        \label{fig:rebuttal_qual}
    \end{figure*}
    
    We depict A2M interaction that represents the connections of appearance-question and question-motion as shown in \figref{fig:rebuttal_qual}.
    Note that we depict the clips as frames sampled at each clip for visibility.
    The frames and words are placed according to temporal and word orders, and corresponding clips of each word are placed regardless of temporal order.
    Although cross-modal interactions are fully-connected, we display the connection with the largest value in each interaction matrix, such that two connected nodes are associated with the maximum interaction value.
    For example, the first frame is associated with the word "man`` by the interaction value of $0.15$, and the word "man`` is related to the forth clip by the interaction value of $0.33$.
    When all the nodes are connected with the equal weights, the values for appearance-question and question-motion interactions are $0.09$ and $0.125$, respectively.


{\small
\bibliographystyle{ieee_fullname}
\bibliography{egbib}
}